\documentclass{esannV2}
\usepackage[latin1]{inputenc}
\usepackage{amssymb,amsmath,array}
\usepackage{multirow} % to allow for cells covering more than 1 row in tables
\usepackage{graphicx}
\usepackage{subfigure}
\usepackage{tabularx} % equal-width columns in tables
\usepackage{hyperref}
\usepackage{arydshln}
\usepackage{url}

\begin{document}
%style file for ESANN manuscripts
\title{Inductive learning for product assortment graph completion}

%***********************************************************************
% AUTHORS INFORMATION AREA
%***********************************************************************
\author{H. Duki\'c$^1$, G. Deligiorgis$^2$, P. Sepe$^1$, D. Bacciu$^1$, and M. Trincavelli$^2$
%
% Optional short acknowledgment: remove next line if non-needed
%\thanks{This is an optional funding source acknowledgement.}
%
% DO NOT MODIFY THE FOLLOWING '\vspace' ARGUMENT
\vspace{.3cm}\\
%
% Addresses and institutions (remove "1- " in case of a single institution)
1- Universit\'a di Pisa - Dipartimento di Informatica \\
Largo B. Pontecorvo 3, 56127 Pisa  - Italy
%
% Remove the next three lines in case of a single institution
\vspace{.1cm}\\
2- H\&M Group  - Business Tech, AI, Analytics, and Data \\
M\"aster Samuelsgatan 46A - Sweden\\
}
%***********************************************************************
% END OF AUTHORS INFORMATION AREA
%***********************************************************************

\maketitle

\begin{abstract}
Global retailers have assortments that contain hundreds of thousands of products that can be linked by several types of relationships like style compatibility, "bought together", "watched together", etc. Graphs are a natural representation for assortments, where products are nodes and relations are edges.
Relations like style compatibility are often produced by a manual process and therefore do not cover uniformly the whole graph. We propose to use inductive learning to enhance a graph encoding style compatibility of a fashion assortment, leveraging rich node information comprising textual descriptions and visual data. Then, we show how the proposed graph enhancement improves substantially the performance on transductive tasks with a minor impact on graph sparsity.
\end{abstract}

\setlength{\extrarowheight}{4pt}

\section{Introduction}
\noindent

Fashion data are interesting for research because of their polymorphism and the complexity of the relations that can be defined among them, i.e. compatibility, transactional, similarity, substitution, etc.
Fashion items are considered compatible if they can be worn simultaneously, meaning that the clothing items are part of an outfit. 
Our work develops on fashion data assembled by H\&M. In this context, the compatibility of fashion items is manually determined by experts on item pairs.
The assortment, however, is composed by tens of thousands articles and the number of pairs grows quadratically with the number of articles, making exhaustive manual labelling highly impractical. Furthermore, when new products enter in the assortment, they stay disconnected for a rather long period. 
The lack of exhaustive indications of item compatibility can considerably impact the performance of recommendation systems that are leveraging such information to provide personalized and style-coherent advice to customers. Motivated by this, we tackle the problem of augmenting such sparse item compatibility information with newly discovered compatibility relationships. 
Existing works have addressed the problem with recurrent models~\cite{han2017learning, nakamura2018outfit, monti2017geometric} or with  contrastive learning \cite{vasileva2018learning, tan2019learning}. Our approach, instead, leverages inductive learning on graphs \cite{bacciu2020gentle}. Inductive link prediction, as opposed to transductive link prediction that assumes all nodes to be present at training time, aims at predicting links for new, unobserved, nodes. However, inductive link prediction usually obtains a lower performance on existing nodes. The method we propose is using inductive link prediction to enrich the graph with new links and then train a transductive model on the new graph to maximize the link prediction performance, getting in this way the best of both worlds. Following \cite{cucurull2019context}, we represent items as nodes and compatibility as edges of the graph, together with their associated information (node and edge labels).
In particular, our items are bound to rich textual and visual information, for which we define an appropriate encoding as node features. We then put forward an inductive learning approach based on the DEAL model \cite{hao2020inductive}, that has been extended to exploit the richness of the multimodal node features available in our industrial case study. As a first result, we show how these features positively contribute to relationship inference. The trained inductive model is then applied to produce an enriched graph for a second transductive task, modelling clothing pairing suggestions as a link prediction problem. The empirical analysis shows that the enriched graph yields to substantially improved link prediction performance over the original graph, at the cost of a minor decrease in graph sparsity. This second result is particularly interesting as it shows the effectiveness and efficiency of a pipeline of inductive-transductive methods when dealing with predictive tasks over large-scale sparse graphs.

\section{Inductive-Transductive Graph Processing Pipeline}

\noindent

For inductive learning, we consider DEAL \cite{hao2020inductive},  an architecture leveraging two encoders, an attribute-oriented encoder $H_a$ and a structure-oriented encoder $H_s$, as well as an alignment mechanism. The aim of the attribute-oriented encoder is to project a node's feature vector from the high-dimensional feature space into a low-dimensional embedding space, while the structure-oriented encoder generates an embedding vector of the node, by considering only the structural information of the graph (no node features). 
If two graph nodes are connected (positive samples), then their $H_a$ and $H_s$ embedding vectors will have high similarity. To this end, we measure  similarity of the embedding vectors by  cosine similarity. 
A Tight Alignment mechanism \cite{hao2020inductive} is used to maximize the similarity between the embedding vectors produced by both $H_a$ and $H_s$ for each node. Both encoders are updated during the training process and the embeddings are kept aligned.
The attribute-oriented encoder can be realized by MLP or GCN-like modules \cite{hao2020inductive}.
In our implementation we use the personalized ranking loss in \cite{hao2020inductive}. 
Transductive learning is implemented with consolidated deep graph networks. In particular, in our empirical analysis, we confront the performance of three popular methods that well represent three different families of neural approaches for graphs that are Graph SAGE \cite{hamilton2017inductive} model, GCN \cite{kipf2016semi} and GAT \cite{velivckovic2017graph}.
As anticipated, the focus of this work is to propose and assess inductive learning as a preliminary step to improve transductive task performance on sparsely connected, large-scale graphs. The process comprises a first step where the DEAL \cite{hao2020inductive} inductive model is trained and the best performing model (in validation) is selected.
% , with respect to the validation loss. 
As a second step, we run the best inductive model selected on the original graph to enrich it with the introduction of new edges. For this second step, we define two thresholds: the maximum node degree and a probability of link existence among the nodes. As a final step, we train the transductive models on the new structure. 
In addition to the pipeline above, we extend DEAL \cite{hao2020inductive} to work on textual, visual, or concatenation of textual and visual features, instead of the tabular features used in \cite{hao2020inductive}. The embedding of the textual and visual information attached to each product in our case study has been obtained by a BERT model \cite{devlin2018bert} pre-trained on English Wikipedia and by a ResNet512 \cite{he2016deep} pre-trained on ImageNet \cite{deng2009imagenet}, respectively.

\section{Item Compatibility Graph}

\noindent

Our study considers two novel industrial proprietary datasets provided by H\&M, where each node is associated with a fashion item and the presence of an edge between two nodes denotes style compatibility between the items. These graphs have been compiled from a list of pairwise fashion item compatibility statements compiled by H\&M domain experts. This information has been used to build two separate graphs, one for Men and one for Women clothing. Both contain a large number of products, represented by an image, text description, colour, and other tabular data. These fashion graphs have been assembled specifically for this work and this is the first graph-based predictive analysis being performed on such data. We complement our analysis on proprietary data with a publicly available dataset, the Computers network \cite{mcauley2015image}, with structural characteristics that are akin to those of our industrial use case.
All networks have been represented using the Open Graph Benchmark (OGB)~\cite{hu2020open} format, and the relevant properties of the aforementioned datasets are given in Table~\ref{tab:graph_summarising}. The challenging aspect shared by all datasets is the high level of edge sparsity, nearing 100\%, and the nontrivial proportion of disconnected nodes (i.e. with zero degree). The latter is particularly true for the fashion data. This is the key motivational aspect for our approach, as we would like to be able to enrich the graph edges by inductive learning before fitting the target transductive task to the data. With respect to this, Table~\ref{tab:graph_summarising} already reports an anticipation of the results of the inductive enrichment of the graph (marked in bold). One can clearly see a considerable drop in disconnected nodes, with a minor change to the edge sparsity.
From a model selection perspective, we split differently the graph adjacency matrix $\textbf{A}$ depending on the task to be performed (inductive or transductive link prediction). In particular, for the inductive link prediction task, we needed to assure that one or both nodes, seen during the training process are not seen during the evaluation process. For this reason, $\textbf{A}$ is split on a node basis. For the transductive task, instead, we partition the network on an edge basis. It is important to mention that negative training edges are sampled uniformly during the training phase, while the validation ones are sampled in advance and are kept fixed for the duration of the model assessment. After the final inductive model is chosen, we set different thresholds for the maximum node degree and a link existence probability. In the case of graph for Men, graph for Women and Computers~\cite{mcauley2015image}, the thresholds for maximum node degree are set to 5, 2, and 20, respectively, while the thresholds for the probability of link existence are set to 0.85, 0.99 and 0.60, respectively.

\begin{table}[ht]
    \begin{center}
    \centering
    \resizebox{\textwidth}{!}{%
            \begin{tabular}{  l | c c| c c |c c } 

            \textbf{Graphs' properties} & 
            \textbf{Men} & \textbf{Men\textsubscript{*}} &
            \textbf{Women} & \textbf{Women\textsubscript{*}} &
            \textbf{Computers} & \textbf{Computers\textsubscript{*}} \\
            \hline
            
            Number of nodes &
            22912 & 22912 & 
            57447 & 57447 &
            13752 & 13752 \\

            Number of zero-degree nodes & 
            5420 & \textbf{16} & 
            11720 & \textbf{2811} & 
            281 & \textbf{0} \\
            
            Percentage of zero degree nodes & 
            23.65\% & \textbf{0.06\%} & 
            20.40\% & \textbf{4.89\%} & 
            2.04\% & \textbf{0.00\%} \\
          
            Number of edges & 
            290514 & \textbf{4265230} & 
            642090 & \textbf{4013554} & 
            491722 & \textbf{3461572}  \\ 
          
            Sparsity & 
            99.94\% & \textbf{99.18\%} &
            99.98\% & \textbf{99.87\%} &
            99.73\% & \textbf{98.16\%} \\
            \hline
        \end{tabular}
        }
        \caption{Dataset summarization: with * we denote enriched graphs.}
        \label{tab:graph_summarising}    
    \end{center}
\end{table}

\noindent

\section{Experiments \& Results}

\noindent
The data used for training the baselines and our proposed method are images and text descriptions of H\&M's assortments. For each dataset, configuration, and task, hyperparameter selection has been performed by using Optuna \cite{akiba2019optuna}, an hyperparameter optimization software framework. For the inductive link prediction task, we trained DEAL-based~\cite{hao2020inductive} models with different architectures and configurations. The results are reported in 
Table~\ref{tab:deal_inductive_grap_for_men_and_women}. In particular, we consider two  attribute-oriented encoder mechanisms: an MLP and a trainable Embedding layer~\cite{Embedding:2019}. The performances for both inductive (Table~\ref{tab:deal_inductive_grap_for_men_and_women}) and transductive (Table~\ref{tab:transductive_results_graph_for_men_and_women}) link prediction are highly improved when using visual or concatenation of visual and textual node features. The best performances on the graph for Men and graph for Women achieved DEAL\textsubscript{MLP}, while  DEAL\textsubscript{EMB} performed better in the case of Computers~\cite{mcauley2015image} for two out of three metrics. The best performing configuration for each graph is used to perform graph enrichment for the successive transductive analysis. 

\begin{table}[htbp!]
    \centering
    \resizebox{\textwidth}{!}{%
        \begin{tabular}{c l| c c c | c c c | c c c}
            &
            \multirow{3}{*}{\textbf{Model}} & 
            \multicolumn{3}{c|}{\textbf{Accuracy}} &
            \multicolumn{3}{c|}{\textbf{ROC - AUC}} &
            \multicolumn{3}{c}{\textbf{AP}} \\ %\cline{2-10}  
            & &  
            \textbf{Text} & \textbf{Image} & \textbf{Text+Image} & 
            \textbf{Text} & \textbf{Image} & \textbf{Text+Image} & 
            \textbf{Text} & \textbf{Image} & \textbf{Text+Image} \\
            
            \hline 
            
            \multirow{2}{*}{\rotatebox[origin=c]{90}{\textbf{Men}}} &
            DEAL\textsubscript{EMB} & 
            0.6432 & 0.7177 &  0.7287 &
            0.7117 & 0.8010 & 0.8208 &
            0.6966 & 0.7463 & 0.7746 \\ [.2em]
            
            &
            \textbf{DEAL\textsubscript{MLP}} & 
            0.7083 & 0.7432 & \textbf{0.7551} &
            0.7728 & 0.7999 & \textbf{0.8370} &
            0.7459 & 0.7438 & \textbf{0.7987} \\ [.4em]
            \hline
            
            \multirow{2}{*}{\rotatebox[origin=c]{90}{\textbf{Women}}} &
            DEAL\textsubscript{EMB} & 
            0.6573 & 0.7258 & 0.7361 &
            0.7349 & 0.8183 & 0.8413 &
            0.7231 & 0.7812 & 0.8088 \\ [.2em]
            
            &
            \textbf{DEAL\textsubscript{MLP}} & 
            0.7276 & 0.7718 & \textbf{0.7851} &
            0.8064 & 0.8402 & \textbf{0.8622} &
            0.7750 & 0.7963 & \textbf{0.8223} \\ [.4em]
            \hline \hline 
            
            \multirow{2}{*}{\rotatebox[origin=c]{90}{\textbf{Comp.}}} &
            \textbf{DEAL\textsubscript{EMB}} & 
            \multicolumn{3}{c|}{0.7425} &
            \multicolumn{3}{c|}{\textbf{0.9072}} &
            \multicolumn{3}{c}{\textbf{0.9019}} \\ [.2em]
            
            &
            DEAL\textsubscript{MLP} & 
            \multicolumn{3}{c|}{\textbf{0.7606}} &
            \multicolumn{3}{c|}{0.8808} &
            \multicolumn{3}{c}{0.8624} \\ [.2em]
            \hline 
            
        \end{tabular}
    }
    \caption{The results of inductive link prediction on Graph For Men, Women, and Computers.}
    \label{tab:deal_inductive_grap_for_men_and_women}
\end{table}

In Table~\ref{tab:transductive_results_graph_for_men_and_women} we can see how the enriched graph improves substantially the link prediction performance for all three different types of GNN considered (SAGE, GAT, and GCN), with respect to the metrics we considered, namely Accuracy, Receiver Operating Characteristic (ROC) Area Under Curve (AUC) and Average Precision (AP).  This improvement in performance may be interpreted by the fact that the enrichment process effectively completes the original graph, making the patterns more regular, general, and therefore easier to learn.

\begin{table}[htbp!]
    \centering
    \resizebox{\textwidth}{!}{%
        \begin{tabular}{c l| c c c | c c c | c c c}
        &
        %\hline
            \multirow{3}{*}{\textbf{Model}} & 
            \multicolumn{3}{c|}{\textbf{Accuracy}} &
            \multicolumn{3}{c|}{\textbf{ROC - AUC}} &
            \multicolumn{3}{c}{\textbf{AP}} \\ 
             &  & 
            \textbf{Text} & \textbf{Image} & \textbf{Text + Image} & 
            \textbf{Text} & \textbf{Image} & \textbf{Text + Image} & 
            \textbf{Text} & \textbf{Image} & \textbf{Text + Image} \\ \hline 
            
            \multirow{6}{*}{\rotatebox[origin=c]{90}{\textbf{Men}}} & 
            SAGE &
            0.9133 &
            0.9256 & 
            0.9323 &
            0.9646 &
            0.9710 & 
            0.9742 &
            0.9563 & 
            0.9649 &
            0.9674 \\ 
            
            &
            \textbf{SAGE\textsubscript{*}} & 
            \textbf{0.9706} &
            \textbf{0.9740} &
            \textbf{0.9736} &
            \textbf{0.9932} &
            \textbf{0.9939} &
            \textbf{0.9944} &
            \textbf{0.9900} &
            \textbf{0.9907} &
            \textbf{0.9915}  \\ 
            \cdashline{2-11}
            
            &
            GCN & 
            0.9180 &
            0.9206 &
            0.9323 &
            0.9687 &
            0.9718 &
            0.9757 &
            0.9649 &
            0.9682 & 
            0.9737 \\ 
            
            &
            \textbf{GCN\textsubscript{*}} & 
            \textbf{0.9704} & 
            \textbf{0.9751} &   
            \textbf{0.9772} &
            \textbf{0.9939} & 
            \textbf{0.9951} & 
            \textbf{0.9956} &
            \textbf{0.9923} & 
            \textbf{0.9935} & 
            \textbf{0.9945} \\ 
            \cdashline{2-11} 
            
            &
            GAT & 
            0.9024 & 
            0.9133 &
            0.9154 &
            0.9565 & 
            0.9658 &
            0.9649 &
            0.9483 & 
            0.9604 &
            0.9588 \\
            
            &
            \textbf{GAT\textsubscript{*}} & 
            \textbf{0.9720} & 
            \textbf{0.9715} & 
            \textbf{0.9750} &
            \textbf{0.9940} &
            \textbf{0.9934} & 
            \textbf{0.9941} &
            \textbf{0.9913} &
            \textbf{0.9889} & 
            \textbf{0.9907} \\ 
            \hline
            
            \multirow{6}{*}{\rotatebox[origin=c]{90}{\textbf{Women}}} & 
            SAGE & 
            0.9214 &
            0.9402 & 
            0.9392 &
            0.9678 & 
            0.9788 & 
            0.9790 &
            0.9627 & 
            0.9750 & 
            0.9756  \\ 
            
            &
            \textbf{SAGE\textsubscript{*}} & 
            \textbf{0.9700} & 
            \textbf{0.9717} & 
            \textbf{0.9711} &
            \textbf{0.9941} & 
            \textbf{0.9946} & 
            \textbf{0.9950} &
            \textbf{0.9932} & 
            \textbf{0.9929} & 
            \textbf{0.9943}  \\ 
            \cdashline{2-11}
            
            &
            GCN & 
            0.9170 & 
            0.9401 &  
            0.9462 &
            0.9695 & 
            0.9813 & 
            0.9839 &
            0.9683 & 
            0.9807 & 
            0.9834 \\ 
            
            &
            \textbf{GCN\textsubscript{*}} & 
            \textbf{0.9690} & 
            \textbf{0.9722} &  
            \textbf{0.9745} &
            \textbf{0.9932} & 
            \textbf{0.9952} & 
            \textbf{0.9951} &
            \textbf{0.9933} & 
            \textbf{0.9947} & 
            \textbf{0.9953} \\ 
            \cdashline{2-11}
            
            &
            GAT & 
            0.9238 & 
            0.9360 & 
            0.9344 &
            0.9695 & 
            0.9759 & 
            0.9752 &
            0.9703 & 
            0.9735 & 
            0.9728\\ 
            
            &
            \textbf{GAT\textsubscript{*}} & 
            \textbf{0.9722} & 
            \textbf{0.9756} & 
            \textbf{0.9748} &
            \textbf{0.9935} & 
            \textbf{0.9946} & 
            \textbf{0.9948} &
            \textbf{0.9916} & 
            \textbf{0.9921} & 
            \textbf{0.9929}\\ 
            \hline \hline
            
            \multirow{6}{*}{\rotatebox[origin=c]{90}{\textbf{Computers}}} & 
            SAGE & 
            \multicolumn{3}{c|}{0.9457} &
            \multicolumn{3}{c|}{0.9832} & 
            \multicolumn{3}{c}{0.9788} \\ 
            
            &
            \textbf{SAGE\textsubscript{*}} & 
            \multicolumn{3}{c|}{\textbf{0.9692}} & 
            \multicolumn{3}{c|}{\textbf{0.9935}} & 
            \multicolumn{3}{c}{\textbf{0.9904}} \\ 
            \cdashline{2-11}
            
            &
            GCN & 
            \multicolumn{3}{c|}{0.9421} &
            \multicolumn{3}{c|}{0.9835} & 
            \multicolumn{3}{c}{0.9818} \\ 
            
            &
            \textbf{GCN\textsubscript{*}} & 
            \multicolumn{3}{c|}{\textbf{0.9647}} & 
            \multicolumn{3}{c|}{\textbf{0.9928}} & 
            \multicolumn{3}{c}{\textbf{0.9910}} \\
            \cdashline{2-11}
            
            &
            GAT & 
            \multicolumn{3}{c|}{0.9364} &
            \multicolumn{3}{c|}{0.9770} & 
            \multicolumn{3}{c}{0.9706} \\ 
            
            &
            \textbf{GAT\textsubscript{*}} & 
            \multicolumn{3}{c|}{\textbf{0.9669}} & 
            \multicolumn{3}{c|}{\textbf{0.9919}} & 
            \multicolumn{3}{c}{\textbf{0.9864}} \\
            \hline
            
        \end{tabular}
        }
    \caption{The results of transductive link prediction on Graph For Men, Women, and Computers. With * are denoted the results of the models after performing the graph enrichment.}
    \label{tab:transductive_results_graph_for_men_and_women}
\end{table}

\section{Conclusions}

\noindent

We proposed an inductive learning approach for completing sparse graphs describing item compatibility information, and we have applied our method to both a publicly available benchmark as well as to a novel industrial use case based on the product assortment of a global fashion retailer. The proposed approach consists of two steps. First, we learn an inductive learning model that we use to generate new links for those nodes of the graph that are disconnected or sparsely connected. We then train a transductive model using the enriched graph, showing that we achieve increased link prediction performance. Our hypothesis is that the inductive learning model manages to learn the patterns of the connected nodes and transfer them to the sparsely connected nodes, making the structure of the graph more regular. This makes sense since we know from the process generating the connections in the graph, which is manual and labor intensive, that many possible connections are missing in the original graph.

Future works will study more thoroughly the graph enrichment step, that in this work has been carried out with a very simple methodology, by selecting the nodes with a maximum number of neighbours and a threshold for the inductive probability prediction. A more principled approach could give further performance improvements.

% ****************************************************************************
% BIBLIOGRAPHY AREA
% ****************************************************************************

\begin{footnotesize}

\bibliographystyle{unsrt}
\bibliography{the_bibliography}

\end{footnotesize}

% ****************************************************************************
% END OF BIBLIOGRAPHY AREA
% ****************************************************************************

\end{document}